\documentclass{article}

\usepackage{microtype}
\usepackage{graphicx}
\usepackage{subcaption}
\usepackage{booktabs} % for professional tables

\usepackage{hyperref}

\usepackage[preprint]{icml2026}

\usepackage{amsmath}
\usepackage{amssymb}
\usepackage{mathtools}
\usepackage{amsthm}

\usepackage[capitalize,noabbrev]{cleveref}

\theoremstyle{plain}

\theoremstyle{definition}

\theoremstyle{remark}

\usepackage[textsize=tiny]{todonotes}

\usepackage{nccmath}
\usepackage{graphicx}
\usepackage{stfloats}

\icmltitlerunning{Multi-Bandwidth Distribution Matching Distillation}

\begin{document}

\twocolumn[
  \icmltitle{Multi-Bandwidth Distribution Matching Distillation: On the Equivalence of Distribution Matching Distillation and Drifting Models}

  % It is OKAY to include author information, even for blind submissions: the
  % style file will automatically remove it for you unless you've provided
  % the [accepted] option to the icml2026 package.

  % List of affiliations: The first argument should be a (short) identifier you
  % will use later to specify author affiliations Academic affiliations
  % should list Department, University, City, Region, Country Industry
  % affiliations should list Company, City, Region, Country

  % You can specify symbols, otherwise they are numbered in order. Ideally, you
  % should not use this facility. Affiliations will be numbered in order of
  % appearance and this is the preferred way.
  \icmlsetsymbol{equal}{*}

  \begin{icmlauthorlist}
    \icmlauthor{Jialin Zhu}{Baidu}
    \icmlauthor{Xing Liu}{Baidu}
    \icmlauthor{Feixiang He}{CSU}
    \icmlauthor{He Wang}{UCL}
  \end{icmlauthorlist}

  \icmlaffiliation{Baidu}{Baidu Inc., China}
  \icmlaffiliation{CSU}{Central South University, China}
  \icmlaffiliation{UCL}{AI Centre, University College London, UK}

  \icmlcorrespondingauthor{Jialin Zhu}{misaliet@outlook.com}
  % \icmlcorrespondingauthor{Firstname2 Lastname2}{first2.last2@www.uk}

  % You may provide any keywords that you find helpful for describing your
  % paper; these are used to populate the "keywords" metadata in the PDF but
  % will not be shown in the document
  \icmlkeywords{Deep Learning}

  \vskip 0.3in
]

% this must go after the closing bracket ] following \twocolumn[ ...

% This command actually creates the footnote in the first column listing the
% affiliations and the copyright notice. The command takes one argument, which
% is text to display at the start of the footnote. The \icmlEqualContribution
% command is standard text for equal contribution. Remove it (just {}) if you
% do not need this facility.

% Use ONE of the following lines. DO NOT remove the command.
% If you have no special notice, KEEP empty braces:
\printAffiliationsAndNotice{}  % no special notice (required even if empty)
% Or, if applicable, use the standard equal contribution text:
% \printAffiliationsAndNotice{\icmlEqualContribution}

\begin{abstract}
Researchers are exploring effective one-step generative model continuously, and, Drifting Models~\cite{deng2026generative}, demonstrate great potential in one-step generation recently.
There are works that reveal the connection between Diffusion \& Flow Style Generative Models (DFSGMs)~\cite{ho2020denoising, song2020denoising, song2020score, lipman2022flow, liu2022flow} and Drifting Models~\cite{li2026long, lai2026unified, turan2026generative}. But no one has yet established a precise correspondence between the Drifting Model and the widely used distillation method- Distribution Matching Distillation (DMD/DMD2)~\cite{yin2024one,yin2024improved} to the best of our knowledge, even though their optimization objective formulas are virtually identical.
In this paper, we prove that by converting the velocity-field / noise-field from the pre-trained DFSGMs into the attraction force field in Drifting Models and estimating the repulsion force field from the generative distribution, training the Drifting Modelis naturally equivalent the Distribution Matching Distillation.
With this equivalent concept, 
we propose an improved method based on DMD from the Drifting Model's perspective- Multi-Bandwidth Distribution Matching Distillation (MBDMD).
\end{abstract}

% ============================================================
% Introduction
% ============================================================

\section{Introduction}

Image generation can be viewed as a distribution mapping problem: given a simple source distribution $p_0$, typically Gaussian noise, and a target data distribution $p_{\mathrm{data}}$, it seek a mapping $h$ such that
\[
h_{\#} p_0 \approx p_{\mathrm{data}}.
\]
Much of the progress in generative modeling over the past decade can be understood as different ways of approaching this mapping. The most direct solution is to learn $h$ itself. GANs~\cite{goodfellow2020generative} follow this idea by training a generator to map noise directly to data, with a discriminator providing the signal that distinguishes the generated distribution from the real one. 
This makes inference a single forward pass, but learning the mapping is difficult: the discriminator changes together with the generator, making optimization require carefully tuned parameters and additional approaches to maintain stability. 
Diffusion and Flow-Matching Style Generative Models (DFSGMs)~\cite{ho2020denoising, song2020denoising, song2020score, lipman2022flow} largely avoid this difficulty by changing the problem. Rather than learning the full mapping from $p_0$ to $p_{\mathrm{data}}$ at once, they break it into a sequence of denoising transformations along a continuous distribution path. In Flow Matching, a time-dependent velocity field specifies how samples should move at each time-step along this path, and the final mapping is obtained by integrating these local updates. This progressive formulation has proved much easier to scale to complex data distributions, but the benefit comes with a structural cost: the learned mapping is no longer executed directly and instead requires iterative inference. 
Recent few-step distillation methods improve inference efficiency while ensuring the quality of the generated results to a certain extent.
Methods such as Distribution Matching Distillation (DMD/DMD2)~\cite{yin2024one, yin2024improved} start from a pretrained Diffusion Model, and train a new student generator to reach the target distribution in only a few steps or even a single step. 
The resulting generator therefore shifts the process from learning local variables to construct trajectories to learning more direct generation by optimizing the distance between distributions.
More recently, Drifting models~\cite{deng2026generative} have introduced a more explicitly designed particles perspective formulation. Instead of adversarial learning or diffusion-based transport and distillation, Drifting Model learns the distribution mapping through explicit attraction toward the target distribution and repulsion from the generated distribution adopt the concept of mean-shift~\cite{cheng1995mean}. Despite this different starting point, it again produces a generator capable of one-step generation. 
Moreover, the Drifting Model has demonstrated immense potential—and not merely in terms of efficiency gains. The high quality of its results has also encouraged further research efforts to build upon it~\cite{han2026one, zhang2026teacher}.
The formulation of the Drifting Model is based on an explicitly human-designed scheme, which raises an interesting question: what is the connection between it and other generative models? Is it merely a variant of some existing method?
There already exist some works that reveal the connection between DFSGMs and Drifting Model from Score-based Generative Model view~\cite{lai2026unified, turan2026generative} or second-order velocity approximation~\cite{li2026long}. In this work, we take a step toward this broader topic. We particularly focus on revealing the connections between DMD and Drifting models, we prove that DMD and Drifting models are theoretically identical computations under distillation of DFSGMs. 
Unlike other works that rely on derivations with asymptotic approximations~\cite{lai2026unified, turan2026generative, li2026long}, we provide a precise correspondence between DMD and Drifting Models with finite-sample analysis.
Based on our derivations, we reformulate the computation in DMD training as that of Drifting Models, and further incorporate the Multi-Bandwidth concept from Drifting Models to improve the training paradigm of DMD. This application of the proposed theorem is named Multi-Bandwidth Distribution Matching Distillation (MBDMD). It enables faster convergence efficiency while achieving better generation quality in DMD training. In addition, the proposed derivation provides natural theoretical explanations for several existing DMD variants from the perspective of Drifting Models.

To summary, our contributions are: (1) We derived the exact correspondence between DMD and Drifting Models training under finite-sample analysis. (2) The proposed derivation provides theoretical bridge from new perspectives for some DMD-related works. (3) Based on our theorem, we proposed an approach that employs multiple bandwidths to improve the training efficiency and results quality of DMD. 
\section{Preliminary}
% \ns{We need to reduce text length in this section.}

{\bf{Diffusion Models}~} construct the transformation from real data distribution to a known noise distribution by progressively perturbing training images until their distribution approaches the tractable noise prior. During the training,  it learns the reverse evolution from noise to real data across noise levels~\cite{sohl2015deep,ho2020denoising}. At inference time, samples drawn from this prior are evolved through the learned reverse process to produce images that approximately follow the data distribution~\cite{song2020score}. This construction is supported by the time-reversal theory of diffusion processes~\cite{anderson1982reverse,song2020score}.
To formalize this, let $x_0\sim p_{\mathrm{data}}$ and consider the forward stochastic differential equation (SDE):
\begin{equation*}
\medmath{
    \mathrm{d}x_t
    = f(x_t,t)\,\mathrm{d}t + g(t)\,\mathrm{d}w_t,
    \qquad t\in[0,1],
}
\end{equation*}
where $f$ is the drift, $g$ is a scalar diffusion coefficient, and $w_t$ is a standard Wiener process. Writing $p_t$ for the marginal density of $x_t$, the forward coefficients are chosen so that $p_1$ is close to a tractable Gaussian prior $p_{\mathrm{prior}}$. The corresponding reverse-time SDE is
\begin{equation*}
\medmath{
    \mathrm{d}x_t
    = \left[
        f(x_t,t)-g(t)^2\nabla_{x_t}\log p_t(x_t)
      \right]\mathrm{d}t
      + g(t)\,\mathrm{d}\bar{w}_t,
}
\end{equation*}
where time runs from $1$ to $0$, $\mathrm{d}t<0$, and $\bar{w}_t$ is a Wiener process in reverse time. The score $\nabla_{x_t}\log p_t(x_t)$ supplies the information about the perturbed data distribution needed to determine this reverse evolution. Learning the reverse process thus reduces to estimating these
scores with a neural network
$s_\theta(x_t,t)\approx\nabla_{x_t}\log p_t(x_t)$.
For generation, this estimate is substituted into the reverse
SDE or the equivalent probability flow ODE.

% \subsection{Flow Matching}
{\bf{Flow Matching}~} (FM)~\cite{lipman2022flow} use the same fundamental concept of forward and reverse process as Diffusion Models. However, it eliminates the dependence of the diffusion process on higher-order randomness. FM's intuition is to directly construct the smoothest continuous deterministic transmission trajectory between the prior noise distribution $p_0(x)$ and the data distribution $p_1(x) = p_{data}(x)$. In practice, the optimal transport is usually utilized to model this trajectory:
\begin{equation*}
\medmath{
% \begin{aligned}
    x_t = t x_1 + (1 - t) x_0, \qquad t \in [0, 1]
}
\label{fm_ot_sche}
% \end{aligned}
\end{equation*}

FM represents the evolutionary ordinary differential equation $\frac{\mathrm{d}x_t}{\mathrm{d}t} = v(x_t, t)$ through a neural velocity field $v_t(x)$. To avoid the difficulty of directly integrating the global marginal vector field $v_t(x)$ in high-dimensional space, FM introduces a conditional prior trajectory. By matching the local velocity field $v_t(x\vert{}x_1) = \frac{x_1 - x}{1-t}$ conditioned on the target data point $x_1$, an unbiased surrogate regression target is established:
\begin{equation*}
\medmath{
    \mathcal{L}_{CFM}(\theta) = \mathbb{E}_{t, x_1 \sim p_{data}, x \sim p_{t\vert{}1}(x\vert{}x_1)} \Vert{} v_\theta(x, t) - v_t(x\vert{}x_1) \Vert{}^2
}
\end{equation*}

Although FM utilizes the optimal transport concept for modeling the trajectory between Gaussian distribution and data distribution. However, learning during training it actually learns the conditional velocity given a data sample instead of real velocity. This makes the learned trajectory curved which also requires a multi-steps sampler for inference.

% \subsection{Distribution Matching Distillation}
{\bf{Distribution Matching Distillation}~} (DMD)~\cite{yin2024one, yin2024improved} is widely used to reduce the cost of generation after training.  
The underlying intuition of DMD is to skip the trajectory integral and forcibly align the generative manifold with few-steps or one-step inference. By distilling the knowledge of high-order complex models into a deterministic generative mapping $G_\theta(z)$, the basic noise $z \sim p_0$ can be translated to the real data space through a single or serveral times forward propagation $x = G_\theta(z)$.
This is achived by minimizing the KL divergence between the generated distribution $p_\theta(x)$ and the target distribution $p_{data}(x)$, the gradient of the parameter update corresponds precisely to the difference in the score matching difference between the two distribution boundary score functions:
\begin{equation*}
% \begin{aligned}
\medmath{
    \nabla_\theta \mathcal{L}_{DMD} = \mathbb{E}_{z \sim p_0} \left[ \left( \nabla_x \log p_\theta(x) - \nabla_x \log p_{data}(x) \right) \frac{d G_\theta}{d \theta} \right]
}
% \end{aligned}
\end{equation*}

In the engineering implementation, the real data score $\nabla_x \log p_{data}(x)$ is approximately provided by the frozen pre-trained multi-step Diffusion Model (teacher model)~\cite{song2020score}, acting as an ``attraction" to pull samples into the manifold. The generated score (fake score) $\nabla_x \log p_\theta(x)$ is provided by a dummy data denoising generator trained simultaneously with the student model to estimate the student model's current generative distribution, providing a ``repulsive force" to avoid model mode collapse. This bidirectional score difference directly guides the mapper $G_\theta$ in non-Euclidean space, thus bypassing the multi-step sampling.

% \subsection{Drifting Models}
{\bf{Drifting Model}~}~\cite{deng2026generative} is a newly proposed single-step generative model. This architecture adopts a mean-shift mechanism~\cite{cheng1995mean} to construct a deterministic, self-correcting dynamical system during training from a particle dynamics perspective. The training process is viewed as a set of particles drifting in a drifting field from the initial Gaussian distribution to the data distribution. Drifting Model constructs a spatial drifting field to guide the generative distribution with 
the attraction force field and the repulsion force field.

The attraction force field $V^+$ ($V_{p, k}(x)$) is defined as the kernel weighted mean shift of the target data distribution $p_1$ under kernel weighting, which dominates the convergence of samples to the true manifold: 
\begin{equation}
\medmath{
    V_{p, k}(x) = \frac{\mathbb{E}{x_1 \sim p_1}\left[(x_1 - x), k(x_1, x)\right]}{\mathbb{E}{x_1 \sim p_1}\left[k(x_1, x)\right]}
}
\label{dm_attr}
\end{equation}

The repulsion field $V^-$ ($V_{q, k}(x)$) is defined as the kernel weighted mean shift of the current generated data distribution $q_{x_t}$, which forces the generated samples to repel each other:
\begin{equation}
\medmath{
    V_{q, k}(x) = \frac{\mathbb{E}{x_t \sim q_{x_t}}\left[(x_t - x), k(x_t, x)\right]}{\mathbb{E}{x_t \sim q_{x_t}}\left[k(x_t, x)\right]}
}
\label{dm_repl}
\end{equation}

The total drift field is a linear superposition of the two potential force fields described above:
\begin{equation}
\medmath{
    V(x) = V_{p, k}(x) - V_{q, k}(x)
}
\label{drifting_field}
\end{equation}

When training a single-step neural network $f_\theta(\varepsilon)$, the model uses the total drift field as the target pseudo-label for geometric displacement. By blocking graph propagation through the stopping gradient operator, the following optimization objective is constructed:
\begin{equation}
\medmath{
    \mathcal{L}(\theta) = \mathbb{E}_\varepsilon \left| f_\theta(\varepsilon) - \text{stopgrad}\left(f_\theta(\varepsilon) + \eta V(f_\theta(\varepsilon))\right) \right|^2
}
\label{drifting_obj}
\end{equation}
\section{Equivalence of Distribution Matching Distillation and Drifting Models}
\label{DMD2DM}
There are works have attempted to unify the perspectives of DFSGMs and Drifting Models.
The Long-Short Flow-Map (LSFM) has proved that Flow Matching is essentially the same with Drifting Models using the second-order approximation near the end point of the trajectory~\cite{li2026long}. Furthermore, researchers has also proved that Drifting Model optimization admits a score-based
formulation~\cite{lai2026unified, turan2026generative}. However, all these methods have only completed non-finite-sample analysis which requires mathematical infinite value approximation. Our derivation, on the contrary, provides an accurate transformation under finite-sample analysis.

In this section, we start by introducing our formulation from the perspective of distilling a pre-trained Flow Matching Model to a one-step Drifting Model. Then, we align our derived formulation with DMD precisely. Finally, we extend this formulation to a more general view and incorporate different kinds of Diffusion Models into our paradigm.

\subsection{Flow Matching View deduction}
\label{FM2DM}

\subsubsection{Velocity to Attraction Force}
\label{v2af}
To distill a one-step student Drifting Model from a multi-step pre-trained teacher Flow Matching model, the essential key is to transfer the velocity-field into the mean-shift attraction force field in Drifting Model.

The core property, anti-symmetric, in Drifting Models does not require a universal fixed kernel function. It only requires the usage of same kernel function for calculating both attraction and repulsion force field. This provides a capable bridge for connecting Drifting Models with pre-trained teacher Flow Matching models, and naturally raises an idea that is to use a special kernel to translate the velocity field into the attraction force field.

Assuming $x_0$ is a sample drawn from Gaussian distribution $\mathcal{N}(0, I)$, the conditional distribution of $x_t$ and the marginal distribution of $p_t(x)$ in Flow Matching can be inferred:
\begin{equation}
\begin{medsize}
\begin{aligned}
    & p(x_t \mid x_1) = \mathcal{N}\left(x_t;\; tx_1,\; (1-t)^2 I\right) \\
    & p_t(x) = \int p_1(x_1)\, \mathcal{N}(x_t;\, tx_1,\, (1-t)^2 I)\, dx_1 \\
\end{aligned}
\end{medsize}
\label{cond_and_marg_dis_p_x_t}
\end{equation}

Expand the conditional distribution in equation~\ref{cond_and_marg_dis_p_x_t} we get:
\begin{equation}
\begin{medsize}
\begin{aligned}
    p(x_t \mid x_1) = C(t) \cdot \exp\left(-\frac{\|x_t - t x_1\|^2}{2(1-t)^2}\right)
\label{cond_p_x_const}
\end{aligned}
\end{medsize}
\end{equation}
Here, $C(t)$ denotes the normalization factor $\frac{1}{\left(2\pi(1-t)^2\right)^{D/2}}$, which is independent to the data $x$ and can be canceled out later.

\textbf{Our design is to use the exponential term in equation~\ref{cond_p_x_const} as the kernel function to calculate the attraction field}:
\begin{equation}
\begin{medsize}
    k_t(y, x) = \exp\left( - \frac{\|ty - x\|^2}{2(1-t)^2} \right) 
\end{medsize}
\label{fm_analytical_kernel}
\end{equation}

Therefore, substituting this kernel function into equation~\ref{cond_p_x_const} yields the following expression for the conditional distribution of $x_t$ in Flow Matching:
\begin{equation}
\begin{medsize}
    p(x_t \mid x_1) = \text{C}(t) \cdot k_t(y, x)
\end{medsize}
\label{cond_p_x_kernel}
\end{equation}

Now substitute equation~\ref{cond_p_x_kernel} to the formula of attraction force field calculation in Drifting Model(equation~\ref{dm_attr}):
\begin{equation}
\begin{medsize}
\begin{aligned}
    V_{p, k}(x) & = \frac{\mathbb{E}_{x_1 \sim p_1}\left[ (x_1 - x) \frac{1}{\text{C}(t)} p(x_t = x \mid x_1) \right]}{\mathbb{E}_{x_1 \sim p_1}\left[ \frac{1}{\text{C}(t)} p(x_t = x \mid x_1) \right]} \\
\label{kernel_p}
\end{aligned}
\end{medsize}
\end{equation}

The $C(t)$ can be can be eliminated, we obtain the attraction force field by further derivation (details are included in the Supplementary Materials~\ref{k2p}):
\begin{equation}
\begin{medsize}
\begin{aligned}
    V_{p, k}(x) & = \frac{\int (x_1 - x) p(x_t = x \mid x_1) p_1(x_1) dx_1}{\int p(x_t = x \mid x_1) p_1(x_1) dx_1} \\
\label{dm_attr_fm_kernel1}
\end{aligned}
\end{medsize}
\end{equation}

According to Bayesian Theory- $\frac{p(x_1, x_t = x)}{p(x_t = x)} = p(x_1 \mid x_t = x)$, substitute this to equation~\ref{dm_attr_fm_kernel1} results in (details are in the Supplementary Materials~\ref{p2e}):
\begin{equation}
\begin{medsize}
\begin{aligned}
    V_{p, k_t}(x) = \mathbb{E}[x_1 \mid x_t = x] - x \\
\label{dm_attr_fm_kernel2}
\end{aligned}
\end{medsize}
\end{equation}

At this point, we have successfully transformed the attraction force field in the Drifting Model into a calculation that is related to the expectation of $x_1$ given $x_t$ ($\mathbb{E}[x_1 \mid x_t = x]$). To $\mathbb{E}[x_1 \mid x_t = x]$ can be infered with Tweedie's Formula~\cite{vincent2011connection}. For Flow Matching, we already have the conditional distribution of $p(x_t \mid x_1)$ (equation~\ref{cond_and_marg_dis_p_x_t}) that is $\mathcal{N}\left(x_t;\; tx_1,\; (1-t)^2 I\right)$. Use Tweedie's Formula, the conditional expectation becomes (deduction is given in the Supplementary Materials~\ref{e2s}):
\begin{equation}
\begin{medsize}
\begin{aligned}
    & \mathbb{E}[x_1 \mid x_t = x] = \frac{x}{t} + \frac{(1-t)^2}{t} s_t(x) \\
\label{expection_ot_fm}
\end{aligned}
\end{medsize}
\end{equation}
where $s_t(x)$ is the score function- $s_t(x) = \nabla_x \log p_t(x)$.

We now have a connection between the conditional expectation and the score function, next substitute equation~\ref{expection_ot_fm} to equation~\ref{dm_attr_fm_kernel2}, we have the relation between the attraction force field in Drifting Models and score in pre-trained Flow Matching Model:
\begin{equation}
\begin{medsize}
% \begin{aligned}
    V_{p, k_t}(x) = \frac{1-t}{t} \left( x + (1-t) s_t(x) \right) 
\label{driting_attraction_ot_fm_2}
% \end{aligned}
\end{medsize}
\end{equation}

The relationship between score function $s_t(x)$ and velocity field $v_t(x)$ in Flow Matching is given in~\cite{lipman2024flow} (we also include detailed deduction in Supplementary Materials~\ref{s2v}):
\begin{equation}
\begin{medsize}
\begin{aligned}
    % & v_t(x) = \frac{1}{t} x + \frac{1 - t}{t} s_t(x) \\
    & s_t(x) = \frac{t\, v_t(x) - x}{1 - t} \\
\label{s_v_uni}
\end{aligned}
\end{medsize}
\end{equation}

Finally, substitute equation~\ref{s_v_uni} to equation~\ref{driting_attraction_ot_fm_2}, we have:
\begin{equation}
\begin{medsize}
    V_{p, k_t}(x) = (1-t) v_t^{real}(x)
\label{fm2dm_attr}
\end{medsize}
\end{equation}
where we successfully transfer the velocity field in a pre-trained Flow Matching Model to  the attraction field in Drifting Model. From this form, we can conclude that, when using the kernel function from the conditional distribution of $x_t$, the corresponding attraction force field in the Drifting Models is simply the product of velocity $v_t(x)$ and remaining time $(1-t)$.

\subsubsection{Repulsion Force and Full Drifting Field}
\label{fm2dm_rep}
Only obtain the attraction force field is not enough to train the student Drifting Model, we also need the corresponding repulsion force field.

The simplest way is to calculate the repulsion directly in the same the velocity space in the pre-trained Flow Matching Model by equation~\ref{dm_repl} from vanilla Drifting Models:
\begin{equation*}
\begin{medsize}
    V_{q, k}(x) = \frac{\mathbb{E}{x_t \sim q_{x_t}}\left[(x_t - x), k_t(x_t, x)\right]}{\mathbb{E}{x_t \sim q_{x_t}}\left[k_t(x_t, x)\right]}
\end{medsize}
\end{equation*}
where $k_t(x_t, x) = \exp\left( - \frac{\|tx_t - x\|^2}{2(1-t)^2} \right) $.

However, as has been proven in the Drifting Model and LSFM~\cite{li2026long}, computing repulsive force field directly in high-dimensional spaces will encounter the intractable dimensionality curse problem for kernel function calculation.

To avoid this, using the Fake Score Model (Fake Flow Matching Model) as the surrogate of repulsion force field in DMD~\cite{yin2024one, yin2024improved} is a practical solution. 

The Fake Flow Matching Model is trained on the student Drifting Model's generative distribution $q_\theta$ —that is, it transmits Gaussian noise $p_0$ to $q_\theta$. The trajectory of Fake Flow Matching Model is:
$x_t^{\text{fake}} = (1-t) \varepsilon + t x_{\text{gen}}$.
The structure of repulsion force field is completely symmetric to the attraction force field deduction that we have in last subsection. So, we can easily get the repulsion force field with the velocity field $v^{\text{fake}}_t(x)$ of Fake Flow Matching Model at time $t$:
\begin{equation}
\begin{medsize}
    V_{q_\theta, k_t}(x) = (1-t)\, v^{\text{fake}}_t(x), \qquad \forall\, x \in \mathbb{R}^D
\end{medsize}
\end{equation}

The full drifting field $f_\theta(\varepsilon)$ of the student Drifting Model can also be obtained with the attraction force field and repulsion force field (we use $v^{\text{real}_t(x)}$ for $v_t(x)$ from pre-trained teacher model):
\begin{equation}
\begin{medsize}
\begin{aligned}
    V(x) & = V_{p, k_t}(x) - V_{q_\theta, k_t}(x) = (1-t)\!\left[v^{\text{real}}_t(x) - v^{\text{fake}}_t(x)\right] \\
\label{drifting_filed_OT_FM}
\end{aligned}
\end{medsize}
\end{equation}

Recall the training objective of Drifting Model:
\begin{equation}
\begin{medsize}
\begin{aligned}
    \mathcal{L}(\theta) = \mathbb{E}_\varepsilon \left\| f_\theta(\varepsilon) - \text{sg}\!\left(f_\theta(\varepsilon) + \eta\, V(f_\theta(\varepsilon))\right) \right\|^2
\label{dm_loss}
\end{aligned}
\end{medsize}
\end{equation}

Substitute Eq~\ref{drifting_filed_OT_FM} to Eq~\ref{dm_loss}, we have the optimization goal for distilling a one-step Drifting Model from a pre-trained Flow Matching Model:
\begin{equation}
\begin{medsize}
\begin{aligned}
    \mathcal{L}(\theta) = &\mathbb{E}_\varepsilon \bigg\| f_\theta(\varepsilon) \\
    &-\text{sg}\!\bigg(f_\theta(\varepsilon) + \eta (1-t)\!\left[v^{\text{real}}_t(f_\theta(\varepsilon)) - v^{\text{fake}}_t(f_\theta(\varepsilon))\right]\bigg) \bigg\|^2
\end{aligned}
\end{medsize}
\end{equation}

\subsection{Connection with DMD}
\label{conn2dmd}

In this section, we show our deduction in last section is essentially the same as DMD/DMD2~\cite{yin2024one, yin2024improved}. DMD calculates the KL divergence between Real Score Model and Fake Score Model.
\begin{equation}
\begin{medsize}
\begin{aligned}
    & D_{KL}(p_{fake}||p_{real}) = \mathbb{E}_{x \sim p_{fake}} \left[ \log \left( \frac{p_{fake}(x)}{p_{real}(x)} \right) \right] \\
    & = \mathbb{E}_{z \sim \mathcal{N}(0, I)} \left[ \log p_{fake}(G_\theta(z)) - \log p_{real}(G_\theta(z)) \right]\\
\end{aligned}
\end{medsize}
\end{equation}

It uses the gradient of KL divergence to optimize the few-step/one-step student model:
\begin{equation}
\begin{medsize}
    \nabla_\theta D_{KL} = \mathbb{E}_{z \sim \mathcal{N}(0, I)} \left[ \left( s_{fake}(G_\theta(z)) - s_{real}(G_\theta(z)) \right) \frac{\partial G_\theta(z)}{\partial \theta} \right]
\end{medsize}
\end{equation}

The score functions of Real Score Model and Fake Score Model are:
\begin{equation}
\begin{medsize}
\begin{aligned}
    s^{real}_t(x_t) = -\frac{x_t - \alpha_t \mu^{real}_t(x_t)}{\sigma_t^2}
\label{real_score}
\end{aligned}
\end{medsize}
\end{equation}
and
\begin{equation}
\begin{medsize}
\begin{aligned}
    s^{fake}_t(x_t) = -\frac{x_t - \alpha_t \mu^{fake}_t(x_t)}{\sigma_t^2}
\label{fake_score}
\end{aligned}
\end{medsize}
\end{equation}

Here the $x_t$ is noised version of student model's output, a Gaussian noise $q_t(x_t|x) \sim \mathcal{N}(\alpha_t x, \sigma_t^2 I)$ is injected into the generated sample. This is to address the issue of scores being $0$ in most regions of high-dimensional manifolds and the Out-of-Distribution (OOD) problem for the pre-trained teacher model. The scores of the real and fake distributions after adding noise are approximated by two independent diffusion models ($\mu_{real}$ and $\mu_{fake}$, aka., the Real Score Model and the Fake Score Model).

Substituting the score of the diffusion model into the gradient formula and introducing a time-weighted term $w_t$, the achieve stable optimization dynamics:
\begin{equation}
\begin{medsize}
    \nabla_\theta D_{KL} \simeq \mathbb{E}_{z, t, x, x_t} \left[ w_t \alpha_t (s^{fake}_t(x_t) - s^{real}_t(x_t)) \frac{d G_\theta}{d \theta} \right]
\end{medsize}
\end{equation}
where $w_t =\frac{\sigma_t^2}{\alpha_t} \frac{CS}{||\mu^{real}_t(x_t) - x||_1}$

The objective of DMD in practical is:
\begin{equation}
\begin{medsize}
\begin{aligned}
    & grad = \mu^{fake}_t(x_t) - \mu^{real}_t(x_t) \\
    & \mathcal{L}_{proxy} = \frac{1}{2} \left\| x - \text{stopgrad}(x - grad) \right\|_2^2 \\
\label{dmd_obj}
\end{aligned}
\end{medsize}
\end{equation}

The equation~\ref{dmd_obj} is similar compared to the Drifting Model training
objective in equation~\ref{drifting_field} and equation~\ref{drifting_obj}. 

If DMD is utilizing a Flow Matching scheme, using the translation between $\varepsilon$-prediction and $v$-prediction from~\cite{lipman2024flow}: $\hat{\epsilon} = x_t + (1-t)v$. Substitute this to equation~\ref{real_score} and equation~\ref{fake_score}, we have:
\begin{equation}
\begin{medsize}
\begin{aligned}
    & s^{real}_t(x_t) = -\frac{x_t + (1-t)v^{real}_t(x_t)}{t} \\
    & s^{fake}_t(x_t) = -\frac{x_t + (1-t)v^{fake}_t(x_t)}{t} \\
\end{aligned}
\end{medsize}
\end{equation}

Substitute the new formulation of real and fake score to:
\begin{equation}
\begin{medsize}
    \nabla_{\theta} D_{KL} \propto s^{fake}_t(x_t) - s^{real}_t(x_t)
\end{medsize}
\end{equation},
we have:
\begin{equation}
\begin{medsize}
\begin{aligned}
    \nabla_{\theta} D_{KL} 
    \propto \frac{1}{t} (1-t) (v^{real}_t(x_t) - v^{fake}_t(x_t)) \\
\end{aligned}
\end{medsize}
\end{equation}
which is basically same as our derivation of equation~\ref{drifting_filed_OT_FM}. The only difference is scaling factor variables.

This is to say the DMD is actually training a Drifting Model with a specific kernel function. The real score is the attraction force field, and the fake score is the repulsion force field in Drifting Models. There are only two differences: the first is the noise-injection in DMD which is unavoidable of OOD issue; the second is the scale factor for the difference of scores. 

\subsection{Unified View deduction}

From section~\ref{FM2DM}, we proved that using specific kernel function can turn the velocity field in Flow Matching to the attraction field in a Student Drifting Model. Further, this kernel has the same form of the exponential term in the expectation of the conditional distribution of $p(x_t \mid x_1)$. We can extend this concept to a more general perspective and also utilize pre-trained Diffusion Models to obtain a student Drifting Model.

We use the unified annotation for Diffusion Models and Flow Matching Models. The $x_0$ is a sample from noise distribution and the $x_1$ is a sample from data distribution. This aligns with the forward process definition of Flow Matching but is the opposite of the definition of Diffusion Model. We reverse the $t$ in the formulation of forward process in Diffusion Models to make the clean data $x_1$ and noise $x_0$. In this way, the unified version of the trajectory is modeled by:
\begin{equation}
\begin{medsize}
    x_t = \alpha_t x_1 + \sigma_t \varepsilon, \qquad \varepsilon \sim \mathcal{N}(0, I)
\end{medsize}
\end{equation},
and the corresponding conditional distribution is:
\begin{equation}
\begin{medsize}
\begin{aligned}
    p(x_t \mid x_1) &= \mathcal{N} \bigl(x_t; \alpha_t x_1,\; \sigma_t^2 I\bigr) \\
    & \propto \exp \left(-\frac{\|x_t - \alpha_t x_1\|^2}{2\sigma_t^2}\right)\\
\label{unified_kernel}
\end{aligned}
\end{medsize}
\end{equation}
The $\alpha_t$ and $\sigma_t$ is the scheduler for the noise perturbation of the data. We give forms for Flow Matching, Diffusion Model Variance Preservation (VP), and Diffusion Model Variance Exploding (VE) in table~\ref{tab:noise_perturbation}.
\begin{table}[h]
    \centering
    \begin{small}
    \begin{tabular}{lcc}
        \toprule
        Method & Signal Scale ($\alpha_t$) & Noise Scale ($\sigma_t$) \\
        \midrule
        Flow Matching & $t$ & $1-t$ \\
        VP Diffusion  & $\alpha_t \in [0,1]$ & $\sqrt{1-\alpha_t^2}$ \\
        VE Diffusion  & $1$ & $\sigma_t \in [0, \sigma_{\max}]$ \\
        \bottomrule
    \end{tabular}
    \caption{Comparison of noise perturbation schedules across different generative frameworks.}
    \label{tab:noise_perturbation}
    \end{small}
\end{table}

We can also apply the Tweedie's Formula again in the unified version of formulation to obtain the unified conditional expectation of equation~\ref{expection_ot_fm}:
\begin{equation}
\begin{medsize}
\begin{aligned}
    & \mathbb{E}[x_1 \mid x_t = x] = \frac{x + \sigma_t^2 s_t(x)}{\alpha_t} \\
\label{expection_general}
\end{aligned}
\end{medsize}
\end{equation}

According to the deduction in section~\ref{FM2DM}, as long as we use the exponential term in the expectation of $p(x_t \mid x_1)$ as the kernel function to calculate the force field, we will always end up with this form no matter what noise scheduler is utilized: $p(x_t \mid x_1) = \text{C}(t) \cdot k_t(y, x)$. Finally, we will always obtain attraction force field as in equation~\ref{dm_attr_fm_kernel2}:
\begin{equation}
\begin{medsize}
\begin{aligned}
    V_{p, k_t}(x) & = \mathbb{E}[x_1 \mid x_t = x] - x \\
\label{dm_attr_general_kernel}
\end{aligned}
\end{medsize}
\end{equation}

Substitute the expectation in equation~\ref{expection_general} to equation~\ref{dm_attr_general_kernel}, we have:
\begin{equation}
\begin{medsize}
\begin{aligned}
    V_{p, k_t}(x) = \frac{(1-\alpha_t) x + \sigma_t^2 s_t(x)}{\alpha_t} \\
\label{dm_attr_general}
\end{aligned}
\end{medsize}
\end{equation}

For Flow Matching Model, we already have the equation~\ref{s_v_uni} and we can get the same result as equation~\ref{fm2dm_attr} with the scheduler where $\alpha_t$ is $t$ and $\sigma_t$ is $(1-t)$.

For Diffusion Model VP, we get the attraction field with the noise-field from teacher model by substitute the VP settings in table~\ref{tab:noise_perturbation}:
\begin{equation}
\begin{medsize}
    V^{+}_{\text{VP}}(x) = \frac{1-\alpha_t}{\alpha_t}\, x \;-\; \frac{\sqrt{1-\alpha_t^2}}{\alpha_t}\, \varepsilon_\theta(x, t)\
\end{medsize}
\end{equation}

For Diffusion Model VE, we get the attraction field also with the noise-field from teacher model by substitute the VE settings in table~\ref{tab:noise_perturbation}:
\begin{equation}
\begin{medsize}
    V^{+}_{\text{VE}}(x) = -\sigma_t\, \varepsilon_\theta(x, t)
\end{medsize}
\end{equation}

The repulsion force field and the full drifting field can be obtained with exactly same methods as in section~\ref{fm2dm_rep}:
\begin{equation}
\begin{medsize}
    V(x) \;=\; V^{+}_t(x) - V^{-}_t(x)
\end{medsize}
\end{equation}
where $V^{-}_t(x)$ has the same form as $V^{+}_t(x)$, except that the noise-field or velocity-field comes from the online updating Fake Score Model.

Finally, we omit the proven of the precise correspondence of our unified formulation with DMD as it is the same we do in section~\ref{conn2dmd}. The difference is the kernel function of different teacher model. In addition to Flow Matching, Diffusion Model VP and Diffusion Model VE, all models that establish trajectories through interpolation in Gaussian distribution and real data distribution can be incorporated into our paradigm.

\subsection{Insight and Connection with DMD-related Studies}
\label{icDMD}

Our formulation naturally provides us with many insights on how to improve DMD-related works if we use the perspective of the Drifting Model. 

One potential lies in how the $t$ is chosen in DMD. From the perspective of DMD, $t$ is used for adding noise and obtaining the marginal score at time $t$ for KL divergence optimization. However, when move the view from DMD to training Drifting Models, this $t$ becomes the bandwidth of the kernel function in many noise schedulers. 

From this perspective, the impact of varying $t$ on the results during DMD distillation can be reinterpreted from a noise-related Signal-to-Noise Ratio (SNR) standpoint to a bandwidth-oriented one. For instance, works such as Flash-DMD~\cite{chen2025flash}, DP-DMD~\cite{wu2026diversity}, and 1.x Distill~\cite{li20261} identified that $t$ significantly influences the SNR in different noise level (high-noise level for global structure and low-noise for local detail). Consequently, they employ different strategies at various noise levels to obtain better optimization signals. 
This allows the exact same result to be derived from a bandwidth perspective.
With a large bandwidth, the attraction tends to point to the center of a large number of samples, and what is learned is general, coarse information from a large number of samples. As the bandwidth gradually decreases, fewer and more precise samples are perceived, and what is learned is detail approaching the ground truth. In addition, works like ADM~\cite{lu2025adversarial} and Phased-DMD~\cite{fan2025phased} improve distillation quality by adjusting the scheduling of $t$ and optimizing the results corresponding to different values of $t$. One uses a Cubic Generator Timestep Schedule to allow more of the $t$ fall into the noisy phase which is the same to utilize more larger bandwidth to perceive more data. The other utilize a Mixture-of-Experts (MoE) architecture to learn results under several different $t$ conditions, which is equivalent to having different Experts handle different perceptual bandwidths.

The above works are basically aimed at solving the learning problem under the selection of $t$ in DMD. Through our formulation, after switching the perspective to Drifting Model, these findings can actually be explained from the perspective of the bandwidth in kernel function computation.

Besides the work mentioned above, another major issue affecting DMD's performance is that the Fake Score Model needs to be updated in real time to track the generative distribution of the student model. This makes training the Fake Score Model difficult and often affects the final quality of the distillation~\cite{ge2025senseflow, huang2026reinforcing}. Our formulation proves that the score estimated by the Fake Score Model is actually the same as the repulsion force field that calculated by a special kernel function. Using the repulsion force field naturally avoids the alignment problem between the generative distributions of the Fake Score Model and the student model. 

However, the biggest problem is that we cannot directly calculate the kernel function in high-dimensional space because of the infamous curse of dimensionality. There exists a ``free, unbiased, and lag-free" estimator for the fake score, but its available bandwidth is too narrow, roughly on the below 0.01 for $t$ according to Effective Sample Size (ESS) calculations. Even so, we demonstrate the feasibility of using the repulsion force field to replace fake scores with low-dimensional data. We use the common 2D checkerboard scenario to implement this, and the results are illustrated in Supplementary Materials~\ref{kc2fsm}. We show that without the curse of dimensionality, it is achievable to address the issue of update lag in the Fake Score Model by using a repulsion force field calculated based on a special kernel function. 
\section{Multi-Bandwidth Distribution Matching Distillation}

To bring advantageous designs from vanilla Drifting Model to DMD by our unified formulation, we analyze the unified version of kernel function (in equation~\ref{unified_kernel}) that from the conditional expectation. We hereinafter refer this kernel function as the Uni kernel.

For the Uni-kernel approach, the variable $\sigma_t$ are theoretically functions of $t$ and represents the bandwidth of force calculation. Typically, $\sigma_t$ decreases as $t$ increasing. When $\sigma_t$ is large, the receptive field calculated by the kernel function is also very large; in this case, it is easy to learn the global structure. As $\sigma_t$ approaches $0$, the bandwidth approaches $0$ too. At this point, the target degenerates into discrete delta spikes on the manifold, causing the model to learn the specific details of the samples. To balance these two extremes, the vanilla Drifting Model employs three different bandwidths to obtain results across varying receptive fields. This motivates our Multi-Bandwidth Distribution Matching Distillation (MBDMD).

The entire MBDMD method can be simply described in one sentence: we divide the entire range of bandwidth values ($t$ values in DMD) into several regions (notated as $K$) and randomly sample a $t$ value as bandwidth from each region which results in various bandwidths at each step of the student model training thereby effectively covering the majority of the bandwidth range.
\section{Experiments and Results}

We show that MBDMD not only improves the quality of the results, but also increase the training efficiency.

The base setting of MBDMD uses $3$ ($K=3$) different bandwidths, with the bandwidth values switching depending on different pre-trained teacher. 

We noticed that choosing a good range of bandwidth values is essential to improve the results quality further. However, the attempt to find a universal method for optimal bandwidth value range by only using a teacher model failed. Therefore, we ultimately chose to use the simplest method to determine the bandwidth sampling distribution, which is to align it with the sampler used during inference for the teacher model.

We follow the vanilla DMD training settings on Diffusion Models with ImageNet dataset at 64 by 64 resolution. Our main experiments were conducted with pure DMD without GAN loss in order to isolate the influence of GAN loss first. Experiments (table~\ref{tab:fid}) show that we achieved better FID results than vanilla DMD's 410k steps with 328k optimization steps. Even considering that we need 3 forward passes per pass (the GPU power consumption of each step is equivalent to 1.15 steps of vanilla DMD), the improvement in training efficiency is substantial. When our experiment reached the same 410k steps, we obtained an FID value of 2.4697.
\begin{table}[H]
    \centering
    \begin{footnotesize}
    \begin{tabular}{lcc}
        \toprule
        Method & steps & FID ($\downarrow$) \\
        \midrule
        DMD & 328k & 2.7238 \\
        DMD & 410k & 2.6432 \\
        MBDMD & 328k & 2.5728 \\
        MBDMD & 410k & \textbf{2.4697} \\
        \bottomrule
    \end{tabular}
    \caption{Comparison of FID of MBDMD and vanilla DMD.}
    \label{tab:fid}
    \end{footnotesize}
\end{table}

To further explore the effective of the settings of components in MBDMD, we did some ablation study experiments (table~\ref{tab:fid_K}). We first tried to use a same $K=3$ but with a i.i.d sampling method. The results are improved but below MBDMD with $K=3$. We also tried using a larger $K$ which is $K=5$, but found that the improvement in results was negligible. The best FID of $K=5$ is below $K=3$ however the iso-flop best FID is winning $K=3$ with around $0.05$. To figure out which setting is better, we divided the training process of $K=3$, $K=5$ and $K=7$ (the GPU power consumption of each step is equivalent to 1.15, 1.29, 1.42 steps of vanilla DMD) into 8 segments according to the number of iso-flop steps to observe the changes in mean FID (shown in table~\ref{tab:fid_K2}). This shows that in most cases $K=3$ is a better choice. Therefore, we chose $K=3$ as the basic setting of MBDMD.

\begin{table}[H]
    \centering
    \begin{scriptsize}
    \begin{tabular}{lcccc}
        \toprule
        Method & $K$ & Sampling & Best FID ($\downarrow$) & Iso-FLOP best FID ($\downarrow$) \\
        \midrule
        DMD & 1 & N/A & 2.6432 & 2.6432 \\
        DMD & 3 & i.i.d & 2.5274 & 2.5702 \\
        MBDMD & 3 & strat & \textbf{2.4697} & 2.5680 \\
        MBDMD & 5 & strat & 2.4819 & \textbf{2.5109} \\
        \bottomrule
    \end{tabular}
    \caption{Comparison of FIDs with different $K$ and sampling method of MBDMD and vanilla DMD.}
    \label{tab:fid_K}
    \end{scriptsize}
\end{table}

\begin{table*}[b]
    \centering
    % \begin{footnotesize}
    % \begin{scriptsize}
    \resizebox{\textwidth}{!}{ 
    \begin{tabular}{lcccccccc}
        \toprule
        % Segment & $K=3$ FID mean/std  & $K=5$ FID mean/std \\
        % \midrule
        Method & Seg 1 avg. FID ($\downarrow$) & Seg 2 avg. FID ($\downarrow$)& Seg 3 avg. FID ($\downarrow$)& Seg 4 avg. FID ($\downarrow$)& Seg 5 avg. FID ($\downarrow$)& Seg 6 avg. FID ($\downarrow$)& Seg 7 avg. FID ($\downarrow$)& Seg 8 avg. FID ($\downarrow$) \\
        % vanilla DMD & 25.535 & 4.526 & 3.795 & 3.467 & 3.331 & 3.183 & 3.113 & 3.045\\
        % vanilla DMD $K=3$ i.i.d & 26.179 & 4.393 & 3.694 & 3.423 & 3.292 & 3.148 & 3.087 & 3.044 \\
        MBDMD $K=3$ & \textbf{25.531} & \textbf{4.285} & \textbf{3.644} & \textbf{3.378} & 3.244 & \textbf{3.110} & 3.124 & \textbf{3.036} \\
        MBDMD $K=5$ & 27.350 & 4.405 & 3.652 & 3.421 & \textbf{3.215} & 3.141 & 3.114 & 3.055 \\
        MBDMD $K=7$ & 29.064 & 4.517 & 3.744 & 3.433 & 3.225 & 3.200 & \textbf{3.096} & 3.063 \\
        \bottomrule
    \end{tabular}
    }
    \caption{Comparison of mean FIDs in every sub-interval of MBDMD with different $K$.}
    \label{tab:fid_K2}
    % \end{footnotesize}
    % \end{scriptsize}
\end{table*}

We also performed training with GAN loss to understand the impact of GAN loss on multi-bandwidth. The results are shown in table~\ref{tab:fid_gan}. When using GAN loss, our method doesn't significantly improve the results with the Imagenet $64\times64$ distillation task. We speculate that this is because the addition of GAN loss mitigates the performance of DMD loss itself. This was actually demonstrated in the ablation study table in the DMD2 paper, where using GAN loss alone without distribution matching loss yielded better FID. 
\begin{table}[H]
    \centering
    \begin{footnotesize}
    \begin{tabular}{lcc}
        \toprule
        Method & steps & FID ($\downarrow$)  \\
        \midrule
        DMD & 157k & 1.5643 \\
        DMD & 200k & 1.5518 \\
        MBDMD & 157k & 1.5304 \\
        MBDMD & 200k & \textbf{1.5304} \\
        \bottomrule
    \end{tabular}
    \caption{Comparison of FID of MBDMD and vanilla DMD with GAN loss.}
    \label{tab:fid_gan}
    \end{footnotesize}
\end{table}

Further, we argue that the performance gain from adding the GAN loss is relatively modest because performance on the ImageNet $64\times64$ distillation task is already approaching its practical limit. To validate this hypothesis, we conducted corresponding distillation experiments on the SDXL model~\cite{podell2024sdxl} (using experimental settings identical to vanilla DMD2). We found that as the model scales up and the data complexity increases, the computational overhead of MBDMD becomes negligible— the cost of a single step is merely 1.03 times that of the vanilla version. Furthermore, even accounting for the overhead introduced by the GAN loss, our method achieves necessary improvements in FID scores (table~\ref{tab:fid_sdxl}). Notably, MBDMD first outperforms vanilla DMD2's 24,000 steps' FID at just 8,000 steps. 
This amply demonstrates that the performance of MBDMD will further improve with more complex data and larger models.

\begin{table}[H]
    \centering
    \begin{scriptsize}
    \begin{tabular}{lcccc}
        \toprule
        Method & Fwd Pass ($\downarrow$) & FID ($\downarrow$) & Patch FID ($\downarrow$) & Clip ($\uparrow$)  \\
        \midrule
        DMD2 & 1 & 19.01 & 26.98 & \textbf{0.336} \\
        MBDMD & 1 & \textbf{18.81} & \textbf{26.86} & 0.335 \\
        \bottomrule
    \end{tabular}
    \caption{Comparison of FID of MBDMD and vanilla DMD.}
    \label{tab:fid_sdxl}
    \end{scriptsize}
\end{table}
\section{Conclusion}
In this paper, we established an exact finite-sample correspondence between DMD and Drifting Models under the distillation frameworks of DFSGMs. This correspondence shows that DMD's distribution-matching computation admits an equivalent attraction--repulsion formulation, revealing a shared theoretical basis beneath their different formulations. Beyond clarifying this connection, the derivation enables the transfer of multi-bandwidth training to DMD, yielding MBDMD and improvements in convergence and generation quality on two image generation datasets. The same theoretical framework also provides an alternative interpretation of several existing DMD variants.

% In the unusual situation where you want a paper to appear in the
% references without citing it in the main text, use \nocite
% \nocite{langley00}

\bibliography{example_paper}
\bibliographystyle{icml2026}

%%%%%%%%%%%%%%%%%%%%%%%%%%%%%%%%%%%%%%%%%%%%%%%%%%%%%%%%%%%%%%%%%%%%%%%%%%%%%%%
%%%%%%%%%%%%%%%%%%%%%%%%%%%%%%%%%%%%%%%%%%%%%%%%%%%%%%%%%%%%%%%%%%%%%%%%%%%%%%%
% APPENDIX
%%%%%%%%%%%%%%%%%%%%%%%%%%%%%%%%%%%%%%%%%%%%%%%%%%%%%%%%%%%%%%%%%%%%%%%%%%%%%%%
%%%%%%%%%%%%%%%%%%%%%%%%%%%%%%%%%%%%%%%%%%%%%%%%%%%%%%%%%%%%%%%%%%%%%%%%%%%%%%%
\newpage
\appendix
\onecolumn
\appendix
% \tableofcontents

\begin{center}
   \section*{\LARGE Supplementary Materials}\label{appendix} 
\end{center}

\section{Detailed Deduction}
\subsection{Kernel Attraction to Probability-based form}
\label{k2p}
We show the deduction of changing the kernel function for attraction force field calculation in Drifting Model to the probability-based computational form. From equation~\ref{kernel_p}, we have:
\begin{equation}
\begin{medsize}
\begin{aligned}
    V_{p, k}(x) & = \frac{\mathbb{E}_{x_1 \sim p_1}\left[ (x_1 - x) \frac{1}{\text{C}(t)} p(x_t = x \mid x_1) \right]}{\mathbb{E}_{x_1 \sim p_1}\left[ \frac{1}{\text{C}(t)} p(x_t = x \mid x_1) \right]} \\
    & = \frac{\frac{1}{\text{C}(t)} \mathbb{E}_{x_1 \sim p_1}\left[ (x_1 - x)  p(x_t = x \mid x_1) \right]}{\frac{1}{\text{C}(t)} \mathbb{E}_{x_1 \sim p_1}\left[ p(x_t = x \mid x_1) \right]} \\
    & = \frac{\mathbb{E}_{x_1 \sim p_1}\left[ (x_1 - x) p(x_t = x \mid x_1) \right]}{\mathbb{E}_{x_1 \sim p_1}\left[ p(x_t = x \mid x_1) \right]} \\
    % & = \frac{\int (x_1 - x) p(x_t = x \mid x_1) p_1(x_1) dx_1}{\int p(x_t = x \mid x_1) p_1(x_1) dx_1} \\
% \label{dm_attr_fm_kernel1}
\end{aligned}
\end{medsize}
\end{equation}

According to the definition of expection- $\mathbb{E}_{z \sim q(z)}[f(z)] \equiv \int f(z) q(z) dz$, we have:
\begin{equation}
\begin{medsize}
\begin{aligned}
    V_{p, k}(x) & = \frac{\mathbb{E}_{x_1 \sim p_1}\left[ (x_1 - x) p(x_t = x \mid x_1) \right]}{\mathbb{E}_{x_1 \sim p_1}\left[ p(x_t = x \mid x_1) \right]} \\
    & = \frac{\int (x_1 - x) p(x_t = x \mid x_1) p_1(x_1) dx_1}{\int p(x_t = x \mid x_1) p_1(x_1) dx_1} \\
\label{dm_attr_fm_kernel1_sm}
\end{aligned}
\end{medsize}
\end{equation}
which is exactly the equation~\ref{dm_attr_fm_kernel1}.

\subsection{Probability-based Attraction form to Expectation-based form}
\label{p2e}

According to Bayesian Theory, we have: 
\begin{equation}
\begin{medsize}
\begin{aligned}
    & p(x_t = x \mid x_1) p_1(x_1) = p(x_1, x_t = x) \\
    & \frac{p(x_1, x_t = x)}{p(x_t = x)} = p(x_1 \mid x_t = x) \\
\end{aligned}
\label{bayesian}
\end{medsize}
\end{equation}

Substitute equation~\ref{bayesian} to equation $V_{p, k}(x) = \frac{\int (x_1 - x) p(x_t = x \mid x_1) p_1(x_1) dx_1}{\int p(x_t = x \mid x_1) p_1(x_1) dx_1}$ (equation~\ref{dm_attr_fm_kernel1}), we obtain:
\begin{equation}
\begin{medsize}
\begin{aligned}
    V_{p, k_t}(x) & = \frac{\int (x_1 - x) p(x_1, x_t = x) dx_1}{\int p(x_1, x_t = x) dx_1} \\
    & = \frac{\int (x_1 - x) p(x_1, x_t = x) dx_1}{p(x_t = x)} \\
    & = \frac{1}{p(x_t = x)} \int (x_1 - x) p(x_1, x_t = x) dx_1 \\
    & = \int (x_1 - x) \frac{p(x_1, x_t = x)}{p(x_t = x)} dx_1 \\
    & = \int (x_1 - x) p(x_1 \mid x_t = x) dx_1 \\
    & = \mathbb{E}[x_1 - x \mid x_t = x] \\
    & = \mathbb{E}[x_1 \mid x_t = x] - \mathbb{E}[x \mid x_t = x] \\
    & = \mathbb{E}[x_1 \mid x_t = x] - x \\
% \label{dm_attr_fm_kernel2}
\end{aligned}
\end{medsize}
\end{equation}
which is the equation~\ref{dm_attr_fm_kernel2}.

\subsection{ Expectation-based Attraction form to score function form}
\label{e2s}

We have the condition distribution of $x_t$ in Flow Matching:
\begin{equation}
\begin{medsize}
\begin{aligned}
    p(x_t \mid x_1) = \mathcal{N}(x_t; t x_1, (1-t)^2 I)
\end{aligned}
\end{medsize}
\end{equation}

The gradient of the log-probability density function of this Gaussian distribution (i.e., the conditional score) is: 
\begin{equation}
\begin{medsize}
\begin{aligned}
    \nabla_{x_t} \log p(x_t \mid x_1) = -\frac{x_t - t x_1}{(1-t)^2}
\end{aligned}
\end{medsize}
\end{equation}

Define the marginal distribution at time $t$ as $p_t(x_t) = \int p(x_t \mid x_1) p_1(x_1) dx_1$. Taking its derivative and expanding using Leibniz's integral rule and Bayes' theorem, we obtain: 
\begin{equation}
\begin{medsize}
\begin{aligned}
    s_t(x_t) & = \nabla_{x_t} \log p_t(x_t) \\
    & = \frac{1}{p_t(x_t)} \nabla_{x_t} \int p(x_t \mid x_1) p_1(x_1) dx_1 \\
    & = \int \frac{\nabla_{x_t} p(x_t \mid x_1)}{p_t(x_t)} p_1(x_1) dx_1 \\
\end{aligned}
\end{medsize}
\end{equation}

Multiply both the numerator and the denominator by $p(x_t \mid x_1)$ to construct the log-gradient:
\begin{equation}
\begin{medsize}
\begin{aligned}
    s_t(x_t) = \int \left[ \nabla_{x_t} \log p(x_t \mid x_1) \right] \frac{p(x_t \mid x_1) p_1(x_1)}{p_t(x_t)} dx_1
\end{aligned}
\end{medsize}
\end{equation}

By Bayes' theorem, the posterior is $p(x_1 \mid x_t) = \frac{p(x_t \mid x_1) p_1(x_1)}{p_t(x_t)}$, so: 
\begin{equation}
\begin{medsize}
\begin{aligned}
    s_t(x_t) = \int \left[ \nabla_{x_t} \log p(x_t \mid x_1) \right] p(x_1 \mid x_t) dx_1 = \mathbb{E}_{x_1 \sim p(x_1 \mid x_t)} \left[ \nabla_{x_t} \log p(x_t \mid x_1) \right]
\end{aligned}
\end{medsize}
\end{equation}

We already have $\nabla_{x_t} \log p(x_t \mid x_1) = -\frac{x_t - t x_1}{(1-t)^2}$, substitute this into last equation we obtain:
\begin{equation}
\begin{medsize}
\begin{aligned}
    s_t(x_t) = \mathbb{E}_{x_1 \sim p(x_1 \mid x_t)} \left[ -\frac{x_t - t x_1}{(1-t)^2} \right]
\end{aligned}
\end{medsize}
\end{equation}

Since both $x_t$ and $t$ are constants with respect to the integration variable $x_1$, we expand the expression using the linearity of expectation: 
\begin{equation}
\begin{medsize}
\begin{aligned}
    s_t(x_t) = -\frac{x_t}{(1-t)^2} + \frac{t}{(1-t)^2} \mathbb{E}[x_1 \mid x_t]
\end{aligned}
\end{medsize}
\end{equation}

Let the current observed value be $x_t = x$. Rearranging the above equation to solve for $\mathbb{E}[x_1 \mid x_t = x]$ yields: 
\begin{equation}
\begin{medsize}
\begin{aligned}
    & \frac{t}{(1-t)^2} \mathbb{E}[x_1 \mid x_t = x] = s_t(x) + \frac{x}{(1-t)^2} \\
    & \mathbb{E}[x_1 \mid x_t = x] = \frac{(1-t)^2}{t} s_t(x) + \frac{x}{t} \\
\end{aligned}
\end{medsize}
\end{equation}

\subsection{Score function to Velocity Field}
\label{s2v}

The trajectory equation for the linear forward process in Flow Matching is given by:
\begin{equation}
\begin{medsize}
\begin{aligned}
    x_t = t x_1 + (1-t) x_0
\end{aligned}
\end{medsize}
\end{equation}

From this, the base noise $x_0$ can be expressed in terms of $x_t$ and $x_1$ via the reverse relation: 
\begin{equation}
\begin{medsize}
\begin{aligned}
    x_0 = \frac{x_t - t x_1}{1-t}
\end{aligned}
\end{medsize}
\end{equation}

Substitute this $x_0$ into the definition of the marginal velocity field $v_t(x_t)$ for Flow Matching: 
\begin{equation}
\begin{medsize}
\begin{aligned}
    v_t(x_t) & = \mathbb{E}\left[x_1 - x_0 \mid x_t\right]\\
    & = \mathbb{E}\left[x_1 - \frac{x_t - t x_1}{1-t} \mid x_t\right]\\
    & = \mathbb{E}\left[\frac{(1-t)x_1 - x_t + t x_1}{1-t} \mid x_t\right] \\
    & = \mathbb{E}\left[\frac{x_1 - x_t}{1-t} \mid x_t\right] \\
    & = \frac{\mathbb{E}[x_1 \mid x_t] - x_t}{1-t} \\
\end{aligned}
\end{medsize}
\end{equation}

Substitute the $\mathbb{E}[x_1 \mid x_t = x] = \frac{x}{t} + \frac{(1-t)^2}{t} s_t(x)$ we have in previous subsection:
\begin{equation}
\begin{medsize}
\begin{aligned}
    v_t(x) & = \frac{1}{1-t} \left( \left[ \frac{x}{t} + \frac{(1-t)^2}{t} s_t(x) \right] - x \right) \\
    & =\frac{1}{1-t} \left( \frac{x - t x}{t} + \frac{(1-t)^2}{t} s_t(x) \right) \\
    & = \frac{1}{1-t} \left( \frac{x(1-t)}{t} + \frac{(1-t)^2}{t} s_t(x) \right) \\
    &= \frac{x}{t} + \frac{1-t}{t} s_t(x) \\
\end{aligned}
\end{medsize}
\end{equation}

We can change the score function to velocity field by reversing the last equation:
\begin{equation}
\begin{medsize}
\begin{aligned}
    s_t(x) = \frac{t \, v_t(x) - x}{1-t}
\end{aligned}
\end{medsize}
\end{equation}

\section{Using Kernel calculation to replace Fake Score Model}
\label{kc2fsm}

In section~\ref{icDMD}, we mentioned that our formulation of the equivalence of Drifting Model and DMD introduce a potential method to resolve the learning lag of online updating for the Fake Score Model in DMD. However, it is not practical in high dimension due to the curse of dimensionality. In order to prove the feasibility, we did an classical 2D toy experiment. 

We used the common 2D checkerboard pattern as the ground truth. First, we trained a Flow Matching Model to serve as the teacher model. Next, we conducted three different experiments, all aimed at obtaining a student model capable of one-step generation. In the first experiment, we directly used the kernel function deducted from with Flow Matching (equation~\ref{fm_analytical_kernel}) to calculate both the attrition and  repulsion forces; the teacher model was not utilized in this case. In the second experiment, we replaced the attraction force with a formulation incorporating the Flow Matching velocity (equation~\ref{fm2dm_attr})—equivalent to the real score in DMD—while still using the Flow Matching kernel (equation~\ref{fm_analytical_kernel}) to calculate the repulsion force. This resolved the issue of update lag in the Fake Score Model. The third experiment was DMD, in which we replaced both the attraction and repulsion forces with formulations incorporating the Flow Matching velocity- specifically, the real score and fake score. The visualization of results are shown in Figure~\ref{kc2fsm}. We can see that all three of these approaches generally yield excellent results.

\begin{figure}[H]
    \centering
    \def\svgwidth{1.0\linewidth}
    % \fontsize{7pt}{9pt}\selectfont
    %% Creator: Inkscape 1.3.2 (091e20e, 2023-11-25), www.inkscape.org
%% PDF/EPS/PS + LaTeX output extension by Johan Engelen, 2010
%% Accompanies image file '2d_toy_exp.pdf' (pdf, eps, ps)
%%
%% To include the image in your LaTeX document, write
%%   \input{<filename>.pdf_tex}
%%  instead of
%%   \includegraphics{<filename>.pdf}
%% To scale the image, write
%%   \def\svgwidth{<desired width>}
%%   \input{<filename>.pdf_tex}
%%  instead of
%%   \includegraphics[width=<desired width>]{<filename>.pdf}
%%
%% Images with a different path to the parent latex file can
%% be accessed with the `import' package (which may need to be
%% installed) using
%%   \usepackage{import}
%% in the preamble, and then including the image with
%%   \import{<path to file>}{<filename>.pdf_tex}
%% Alternatively, one can specify
%%   \graphicspath{{<path to file>/}}
%% 
%% For more information, please see info/svg-inkscape on CTAN:
%%   http://tug.ctan.org/tex-archive/info/svg-inkscape
%%
\begingroup%
  \makeatletter%
  \providecommand\color[2][]{%
    \errmessage{(Inkscape) Color is used for the text in Inkscape, but the package 'color.sty' is not loaded}%
    \renewcommand\color[2][]{}%
  }%
  \providecommand\transparent[1]{%
    \errmessage{(Inkscape) Transparency is used (non-zero) for the text in Inkscape, but the package 'transparent.sty' is not loaded}%
    \renewcommand\transparent[1]{}%
  }%
  \providecommand\rotatebox[2]{#2}%
  \newcommand*\fsize{\dimexpr\f@size pt\relax}%
  \newcommand*\lineheight[1]{\fontsize{\fsize}{#1\fsize}\selectfont}%
  \ifx\svgwidth\undefined%
    \setlength{\unitlength}{1004.94157926bp}%
    \ifx\svgscale\undefined%
      \relax%
    \else%
      \setlength{\unitlength}{\unitlength * \real{\svgscale}}%
    \fi%
  \else%
    \setlength{\unitlength}{\svgwidth}%
  \fi%
  \global\let\svgwidth\undefined%
  \global\let\svgscale\undefined%
  \makeatother%
  \begin{picture}(1,0.26151617)%
    \lineheight{1}%
    \setlength\tabcolsep{0pt}%
    \put(0,0){\includegraphics[width=\unitlength,page=1]{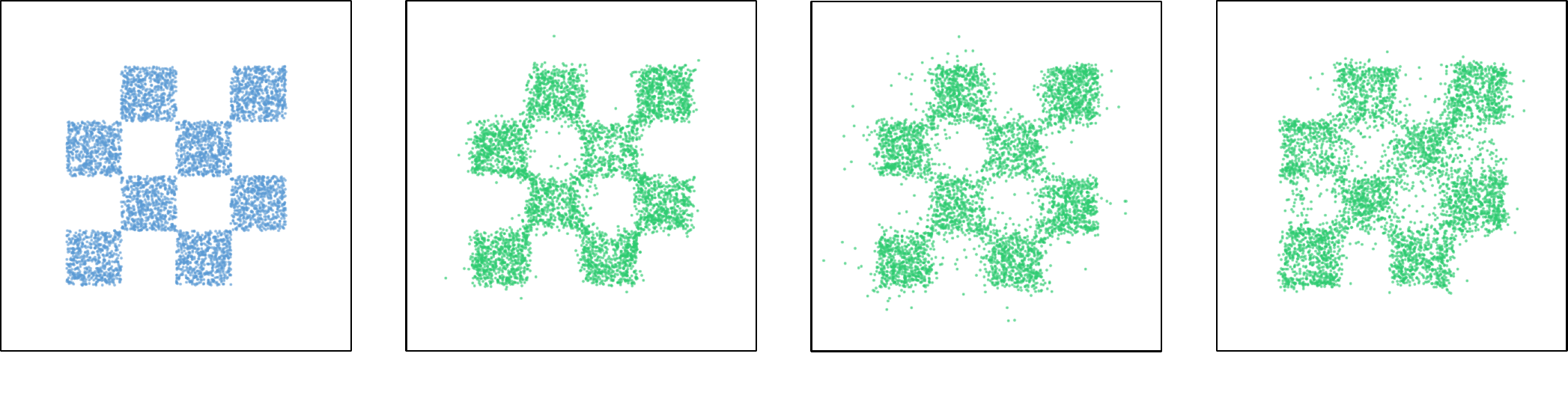}}%
    \put(0.1122451,0.00638444){\makebox(0,0)[t]{\lineheight{1.25}\smash{\begin{tabular}[t]{c}(a)\end{tabular}}}}%
    \put(0.37104662,0.00638444){\makebox(0,0)[t]{\lineheight{1.25}\smash{\begin{tabular}[t]{c}(b)\end{tabular}}}}%
    \put(0.62920836,0.00638444){\makebox(0,0)[t]{\lineheight{1.25}\smash{\begin{tabular}[t]{c}(c)\end{tabular}}}}%
    \put(0.88805316,0.00638444){\makebox(0,0)[t]{\lineheight{1.25}\smash{\begin{tabular}[t]{c}(d)\end{tabular}}}}%
  \end{picture}%
\endgroup%

    \caption{\textbf{A 2D toy experiments to show that Fake Score Model can be replaced by the repulsion kernel function calculation.} In this figure, (a) is the ground-truth; (b) is the results generated by pure kernel calculation for both attraction and repulsion; (c) shows the result by transferring velocity from pretrained Flow Matching Model to calculate attraction force and use kernel calculation for repulsion; (d) is the result from DMD which using velocity from pretrained Flow Matching Model to calculate attraction force and using a online updating Fake Score Model to calculate repulsion force.}
    \label{fig:kc2fsm}
\end{figure}

\end{document}